\documentclass{article}
\usepackage{arxiv}

\usepackage{algorithm}
\usepackage{algpseudocode}
\usepackage{amsmath}
\usepackage[utf8]{inputenc} 
\usepackage[T1]{fontenc}    
\usepackage{url}            
\usepackage{booktabs}       
\usepackage{amsfonts}       
\usepackage{nicefrac}       
\usepackage{microtype}      
\usepackage{lipsum}
\usepackage{amsmath}
\usepackage{amssymb}
\usepackage{graphicx}
\usepackage{times}
\usepackage{geometry}       
\usepackage{float}          
\usepackage{tikz}
\usetikzlibrary{shapes.geometric, arrows, positioning, calc, fit, patterns}

\usepackage{hyperref}
\hypersetup{
    colorlinks=true,
    linkcolor=black,
    filecolor=magenta,      
    urlcolor=blue,
    citecolor=blue,
    pdfpagemode=FullScreen,
}

\title{\textbf{Neuro-Channel Networks: A Multiplication-Free Architecture by Biological Signal Transmission}}

\author{
  \begin{minipage}[t]{0.48\textwidth}
    \centering
    \textbf{Emrah Mete} \\
    Department of Computer Engineering\\
    Yeditepe University\\
    Istanbul, Türkiye \\
    \small \texttt{emrah.mete@std.yeditepe.edu.tr}
  \end{minipage}
  \hfill
  \begin{minipage}[t]{0.48\textwidth}
    \centering
    \textbf{Emin Erkan Korkmaz} \\
    Department of Computer Engineering\\
    Yeditepe University\\
    Istanbul, Türkiye \\
    \small \texttt{ekorkmaz@cse.yeditepe.edu.tr}
  \end{minipage}
}

\date{}

\begin{document}

\maketitle

\begin{abstract}
The rapid proliferation of Deep Learning is increasingly constrained by its heavy reliance on high-performance hardware, particularly Graphics Processing Units (GPUs). These specialized accelerators are not only prohibitively expensive and energy-intensive but also suffer from significant supply scarcity, limiting the ubiquity of Artificial Intelligence (AI) deployment on edge devices. The core of this inefficiency stems from the standard artificial perceptron's dependence on intensive matrix multiplications. However, biological nervous systems achieve unparalleled efficiency without such arithmetic intensity; synaptic signal transmission is regulated by physical ion channel limits and chemical neurotransmitter levels rather than a process that can be analogous to arithmetic multiplication. Inspired by this biological mechanism, we propose \textbf{Neuro-Channel Networks (NCN)}, a novel multiplication-free architecture designed to decouple AI from expensive hardware dependencies. In our model, weights are replaced with ``Channel Widths'' that physically limit the signal magnitude, while a secondary parameter acts as a ``Neurotransmitter'' to regulate \textbf{Signal Transmission} based on sign logic. The forward pass relies exclusively on addition, subtraction, and bitwise operations (minimum, sign), eliminating floating-point multiplication entirely. In this proof-of-concept study, we demonstrate that NCNs can solve non-linearly separable problems like XOR and the Majority function with 100\% accuracy using standard backpropagation, proving their capability to form complex decision boundaries without multiplicative weights. This architecture offers a highly efficient alternative for next-generation neuromorphic hardware, paving the way for running complex models on commodity CPUs or ultra-low-power chips without relying on costly GPU clusters.
\end{abstract}

\section{Introduction}

Deep Neural Networks (DNNs) have achieved remarkable success in tasks ranging from computer vision to natural language processing. However, this performance comes at a substantial computational cost. The fundamental operation underlying nearly all modern DNNs is matrix multiplication. Although effective, floating-point multiplication is computationally expensive and energy intensive, posing a significant bottleneck for the deployment of advanced AI applications on edge devices, IoT sensors, and neuromorphic hardware with strict power constraints~\cite{ref1}.

To address this ``multiplication bottleneck,'' researchers have explored various alternatives. Binary Neural Networks (BNNs) restrict weights to $\{-1, +1\}$ to replace multiplications with bitwise operations~\cite{ref2}, often at the cost of accuracy. More recently, AdderNets~\cite{ref3} proposed replacing dot products with the L1-norm (Manhattan) distance, demonstrating that additions alone can construct models which can achieve convergence on non-linear problems.  However, the performance of the proposed methods is limited compared to the DNN models that are heavily dependent on matrix multiplications. 

Although these approaches successfully reduce computational complexity, they are primarily mathematical approximations rather than biologically plausible mechanisms. In biological nervous systems, no process analogous to  floating-point arithmetic multiplications can be observed in synapses. 
Instead, signal transmission is regulated by physical constraints and chemical modulators. Ion channels act as gating mechanisms that physically limit the magnitude of signal flow (channel saturation), while neurotransmitter levels determine synaptic efficacy~\cite{ref4}.

Inspired by this biological efficiency, we propose a novel architecture: \textbf{Neuro-Channel Networks (NCN)}. NCN is a fully multiplication-free architecture designed to solve non-linear problems using only addition, subtraction, and bitwise logic operations (such as minimum, sign, and multiplexing). Unlike AdderNets~\cite{ref3} which rely on distance metrics, NCN is inspired from the \textbf{Biological Signal Transmission} mechanism. In our model, weights are replaced with physical \textbf{Channel Widths} that clamp the input signal, while a secondary parameter acting as a \textbf{Neurotransmitter} creates a learnable bypass path to regulate signal flow and gradient propagation.

In this work, we present a proof-of-concept for NCNs. Our contributions are as follows:

\begin{enumerate}
    \item We introduce the \textbf{Neuro-Channel Layer}, a processing unit that eliminates multiplication in the forward pass by combining channel-limiting logic with a sign-based transmission mechanism.
    \item We integrate a signal scaling factor based on input dimension to maintain variance stability without complex initialization schemes~\cite{ref7}.
    \item We demonstrate that despite lacking multiplication, NCNs can solve non-linearly separable problems (e.g., XOR gate, Majority function) with 100\% accuracy using standard backpropagation~\cite{ref6}, proving their capability to form complex decision boundaries without multiplicative weights.
\end{enumerate}

This architecture offers a promising new direction for extreme-edge AI hardware where substantial reductions in silicon area and energy consumption can be obtained.

\section{Related Work}

The pursuit of removing computational burden in Deep Neural Networks (DNNs) has led to a chronological evolution of architectures, transitioning from low-precision approximations to arithmetic replacement strategies. The initial wave of efficiency-focused research targeted weight quantization. Courbariaux et al.~\cite{ref2} pioneered \textbf{Binary Neural Networks (BNNs)}, effectively constraining both weights and activations to $\{-1, +1\}$. This radical quantization replaced heavy floating-point multiplications with bitwise XNOR operations and population counts (popcount), theoretically reducing memory usage by $32\times$ and operations by $58\times$ on CPUs. Following this, Rastegari et al. proposed \textbf{XNOR-Net}~\cite{ref9}, which introduced scaling factors to recover some of the accuracy lost during binarization. However, while these approaches offer extreme speedups, they often suffer from significant information loss and training instability due to the discontinuity of the sign function, limiting their expressiveness on complex datasets.

As an intermediate step between strict binarization and full precision, researchers explored replacing multiplication with bit-shift operations. Architectures like \textbf{DeepShift}~\cite{ref10} constrain weights to be powers of 2 (e.g., $2^p$). Since binary shifting is computationally cheaper than multiplication, these networks offer a middle ground with better accuracy than BNNs. Nevertheless, they are still fundamentally based on approximating the multiplicative nature of standard neurons rather than redefining the accumulation mechanism itself, and they often require specialized hardware shifters to realize their full theoretical efficiency.

A significant paradigm shift occurred with the introduction of \textbf{AdderNets} by Chen et al.~\cite{ref3}. Challenging the necessity of the dot product, AdderNets replaced the convolution operation with a similarity measure based on the $L_1$-norm (Manhattan distance), where feature responses are computed via additive operations rather than multiplications. This approach demonstrated that feature extraction using only addition and subtraction can be highly effective, achieving competitive performance with standard CNNs on ResNet-50 benchmarks. From a biological perspective, the reliance on a distance-based similarity metric differs from the signal transmission mechanisms typically associated with synaptic interactions. Moreover, the non-smooth gradient characteristics of the $L_1$ distance require specialized adaptive learning rate strategies to ensure stable and efficient convergence.

Parallel to these structural optimizations, biologically plausible models such as \textbf{Spiking Neural Networks (SNNs)}~\cite{ref11} have attempted to eliminate continuous multiplication by relying on discrete spike events. While highly energy-efficient on neuromorphic hardware, SNNs are notoriously difficult to train due to the non-differentiable nature of spike generation.

Our proposed \textbf{Neuro-Channel Networks (NCN)} occupy a distinct position in this landscape. Unlike BNNs, NCNs do not suffer from the extreme information loss of 1-bit quantization. Unlike AdderNets, our architecture does not rely on mathematical distance metrics; instead, it models the synapse as a flow-control mechanism governed by physical channel limits and neurotransmitter gating. This approach aligns the mathematical operation with biological plausibility and mimicking ion channel saturation while retaining the gradient flow stability, effectively bridging the gap between biological fidelity and deep learning efficiency.

\section{Methodology}

In this section, we formally introduce the \textbf{Neuro-Channel Network (NCN)} architecture. Our primary objective is to construct a neural network model in which the forward propagation phase is completely free of floating-point multiplication operations, relying solely on addition, subtraction, and bitwise logic.

\subsection{The Neuro-Channel Perceptron}

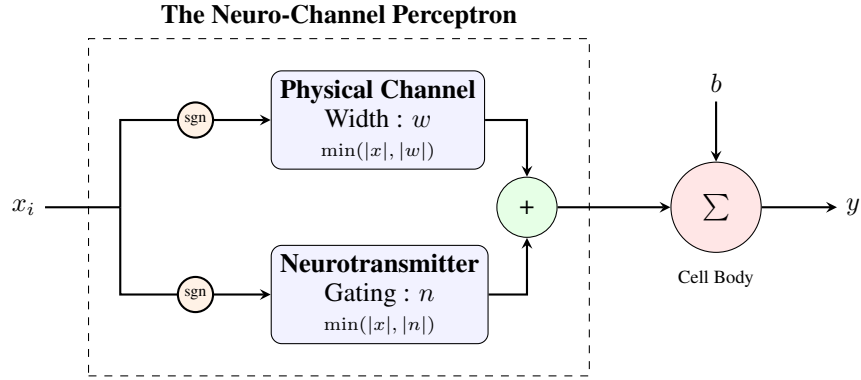
\begin{figure}[htbp]
    \centering
    \begin{tikzpicture}[
        node distance=1.5cm,
        process/.style={draw, rectangle, minimum width=2.5cm, minimum height=1.2cm, align=center, fill=blue!5, rounded corners},
        decision/.style={draw, diamond, aspect=1.5, align=center, fill=yellow!10},
        sum/.style={draw, circle, minimum size=0.8cm, fill=green!10, node distance=2cm},
        io/.style={draw=none, align=center},
        arrow/.style={->, thick, >=stealth},
        sgn_node/.style={circle, draw, fill=orange!10, inner sep=1.5pt, font=\tiny}
    ]

    \node[io] (input) {$x_i$};

    \coordinate[right=1cm of input] (split);

    \node[process, above right=0.5cm and 2cm of split] (channel) {
        \textbf{Physical Channel}\\
        Width : $w$\\
        \scriptsize $\min(|x|, |w|)$
    };

    \node[process, below right=0.5cm and 2cm of split] (neuro) {
        \textbf{Neurotransmitter}\\
         Gating : $n$\\
        \scriptsize $\min(|x|, |n|)$
    };

    \node[sum, right=6cm of input] (plus) {+};

    \node[sum, right=1.5cm of plus, fill=red!10, minimum size=1.2cm] (sigma) {$\sum$};
    \node[below=3pt] at (sigma.south) {\scriptsize Cell Body};

    \node[above=0.8cm of sigma] (bias) {$b$};
    \node[io, right=1cm of sigma] (output) {$y$};

    \draw[thick] (input) -- (split);
    \draw[arrow] (split) |- node[pos=0.75, sgn_node] {sgn} (channel);
    \draw[arrow] (split) |- node[pos=0.75, sgn_node] {sgn} (neuro);
    \draw[arrow] (channel) -| (plus);
    \draw[arrow] (neuro) -| (plus);
    \draw[arrow] (plus) -- (sigma);
    \draw[arrow] (bias) -- (sigma);
    \draw[arrow] (sigma) -- (output);

    \node[draw, dashed, inner sep=12pt, label=above:\textbf{The Neuro-Channel Perceptron}, fit=(channel) (neuro) (plus) (split)] (synapse_box) {};

    \end{tikzpicture}
    
    \caption{Structure of a single \textbf{Neuro-Channel Perceptron}. The input signal $x_i$ is processed through two parallel paths: a \textbf{Physical Channel} path constrained by the weight width $|w|$, and a \textbf{Chemical Bypass} path gated by the neurotransmitter level $|n|$. Both paths preserve the original sign of the input.}
    \label{fig:neuron_diagram}
\end{figure}

Standard artificial perceptrons compute the total sum $in_j$ for some perceptron  $j$ as the weighted sum of inputs using the dot product operation: $in_j = \sum_{i} (w_i \cdot x_i) + b$ ~\cite{ref16} where $x_i$ is the input of unit $i$, $w_i$ is the corresponding weight and $b$ is the bias input. This multiplication operation represents the projection of the input vector onto the weight vector. In contrast, our proposed perceptron models synaptic transmission as a physical flow process regulated by channel constraints. We define the output of a Neuro-Channel perceptron as the aggregation of two parallel mechanisms: the \textbf{Channel Width} (physical limit) and the \textbf{Neurotransmitter Bypass} (chemical regulation).
The structural overview of this synaptic processing unit is illustrated in Figure~\ref{fig:neuron_diagram}.

\subsubsection{Channel Width (Physical Limit)}
Let $x \in \mathbb{R}^d$ be the input vector and $w \in \mathbb{R}^d$ be the weight vector. In standard neural networks, weights act as multiplicative scaling factors ($w \cdot x$). However, in our architecture, $w$ represents the physical \textbf{width} or \textbf{capacity} of the channels. 

The channel width creates a non-linear gating mechanism that limits the magnitude of the input signal based on this channel width $|w|$, while strictly preserving the direction (sign) of the input. Unlike matrix multiplication, which projects the input onto the weight vector, this operation acts as a \textbf{dynamic signal clamper}. This is mathematically defined as:

\begin{equation}
    C(x, w) = \text{sgn}(x) \odot \min(|x|, |w|)
\end{equation}

where $\odot$ denotes element-wise application. 

\textbf{Physical Interpretation:} This mechanism mirrors the saturation behavior of biological synapses. Just as a channel can only allow a maximum flow rate of ions determined by its pore size regardless of the concentration gradient, the Neuro-Channel neuron allows the signal $x$ to pass through only up to the limit defined by $|w|$. If the input signal is weak ($|x| < |w|$), it flows freely; if it is strong ($|x| > |w|$), it saturates at the channel's limit. 

In hardware implementation, this logic does not require floating-point arithmetic. It is performed efficiently as a conditional selection (multiplexer): if the sign bit of $x$ is positive, the circuit transmits $\min(|x|, |w|)$; otherwise, it transmits $-\min(|x|, |w|)$.

\subsubsection{Neurotransmitter Bypass}
In biological synapses, signal transmission is not solely dependent on the physical pore size of the ion channels. Neurotransmitters (chemical messengers) actively modulate synaptic efficacy through slow transmission pathways, allowing signals to propagate even when physical channels might be momentarily constrained~\cite{ref13}.

We model this phenomenon to address a critical optimization challenge known as the ``dead gradient'' problem. In our channel model, if a width parameter becomes very small ($|w| \approx 0$) or significantly smaller than the input ($|w| \ll |x|$), the channel effectively saturates or closes. In a standard hard-clipping scenario, this would block the gradient flow, halting the learning process for that neuron.

To prevent this, we introduce a secondary learnable parameter vector $n \in \mathbb{R}^d$, representing \textbf{Neurotransmitter Levels}. This creates a parallel, chemically-regulated pathway that operates alongside the physical channel:

\begin{equation}
    B(x, n) = \text{sgn}(x) \odot \min(|x|, |n|)
\end{equation}

\textbf{Functional Role:} This mechanism acts as a \textit{learnable leakage} or a biological equivalent of a \textit{residual connection}~\cite{ref14}. Even if the physical channel ($w$) is fully saturated or shut down, the signal can still bypass the restriction through the neurotransmitter path ($n$). This ensures that:
\begin{itemize}
    \item \textbf{Information Flow:} The forward pass preserves signal variance even through narrow channels.
    \item \textbf{Gradient Propagation:} The backward pass always has an active path for error derivatives to flow, ensuring that weights continue to update and escape local minima efficiently.
\end{itemize}

\subsubsection{Total Output and Scaling}
After processing the input signal through the parallel mechanisms of \textit{the physical channel} model (Step 2 in Algorithm~\ref{alg:ncn_forward}) and \textit{Neurotransmitter Bypass} (Step 3), the resulting signal components must be aggregated to form the final activation of the perceptron. This phase models the biological process of \textbf{Somatic Integration} (specifically \textit{spatial summation}), where excitatory and inhibitory postsynaptic potentials are integrated in the cell body (soma)~\cite{ref4}.

The total output $y_j$ is computed by summing the channel output $C(x_i, w_{ji})$ and the neurotransmitter output $B(x_i, n_{ji})$ across all input connections $i=1 \dots d$. However, in deep networks, a direct summation of the $d$ inputs typically causes the signal variance to grow linearly with the input dimension, leading to the well-known ``exploding gradient'' problem. In biological systems, neurons prevent this over-excitation through \textbf{homeostatic plasticity} mechanisms, such as synaptic scaling, which stabilize firing rates against drastic changes in input drive~\cite{ref12}.

To ensure signal stability and mimic this biological regulation without relying on complex weight initialization schemes, we apply a static scaling factor derived from the input dimension $d$. This normalization step corresponds to \textbf{Phase 5} in our algorithm: 

\begin{equation}
    y_j = \left( \frac{\sum_{i=1}^{d} (C(x_i, w_{ji}) + B(x_i, n_{ji}))}{\sqrt{d}} \right) + b_j
\end{equation}

The term $\sqrt{d}$ acts as a normalization factor analogous to biological homeostatic scaling, mathematically serving as a variance preserving factor~\cite{ref8}, aligning with the scaling principles proposed by LeCun et al.~\cite{ref7}.  \textbf{It is important to note the distinction in computational complexity:} Unlike standard artificial neurons where multiplication operations scale linearly with the input dimension $O(d)$ (occurring at every synaptic connection), this scaling is a static scalar operation applied only once per neuron during somatic integration ($O(1)$). Consequently, the high volume synaptic processing remains entirely multiplication-free, and the computational cost of this single global scaling step is negligible compared to the dense matrix multiplications required by deep neural networks.

The complete computational flow, linking these physical and chemical mechanisms to the final output, is detailed in Algorithm~\ref{alg:ncn_forward}.

\begin{algorithm}[H]
\caption{Forward Propagation of Neuro-Channel Layer}
\label{alg:ncn_forward}
\begin{algorithmic}[1]
\Require Input vector $\mathbf{x}$, Weights $\mathbf{W}$, Neurotransmitters $\mathbf{N}$, Bias $\mathbf{b}$
\Ensure Output vector $\mathbf{y}$
\State \textbf{Initialize:} $\mathbf{y} \gets \mathbf{0}$, $d \gets \text{length}(\mathbf{x})$
\For{each neuron $j$ in layer}
    \State $sum\_val \gets 0$
    \For{each input feature $i$}
        \State \textbf{// 1. Compute Magnitudes (No Multiplication)}
        \State $abs\_x \gets |x_i|$
        \State $abs\_w \gets |W_{ji}|$
        \State $abs\_n \gets |N_{ji}|$
        
        \State \textbf{// 2. Physical Channel Model (Gating)}
        \State $limit\_c \gets \min(abs\_x, abs\_w)$
        \If{$\text{sgn}(x_i) == \text{sgn}(W_{ji})$}
            \State $channel\_out \gets \text{sgn}(x_i) \cdot limit\_c$ \Comment{Signal flows freely}
        \Else
            \State $channel\_out \gets -limit\_c$ \Comment{Signal inverted (Inhibitory)}
        \EndIf
        
        \State \textbf{// 3. Neurotransmitter Bypass (Leakage)}
        \State $limit\_n \gets \min(abs\_x, abs\_n)$
        \State $neuro\_out \gets \text{sgn}(x_i) \cdot limit\_n$ \Comment{Bypass follows input sign}
        
        \State \textbf{// 4. Accumulate Signals}
        \State $sum\_val \gets sum\_val + (channel\_out + neuro\_out)$
    \EndFor
    
    \State \textbf{// 5. Somatic Integration \& Normalization}
    \State $y_j \gets (sum\_val / \sqrt{d}) + b_j$ \Comment{Scale to preserve variance}
\EndFor
\State \Return $\mathbf{y}$
\end{algorithmic}
\end{algorithm}

\subsection{Computational Complexity Analysis}
To accurately assess efficiency gains, we compare the theoretical operation count of a standard Feed-Forward Neural Network (FFNN) perceptron against a Neuro-Channel perceptron for an input vector of dimension $d$. We categorize operations into two distinct phases: \textbf{Synaptic Operations}, which scale linearly with input dimension ($O(d)$), and \textbf{Somatic Operations}, which occur once per neuron ($O(1)$).

As illustrated in Table~\ref{tab:complexity}, a standard perceptron is highly dependent on $d$ multiplication operations within the synaptic loop, which constitutes the primary energy bottleneck in deep networks. In contrast, Neuro-Channel Networks (NCN) completely eliminate multiplications at the synaptic level. Although our architecture introduces a scaling operation, it is crucial to observe its location in the computational graph: it is a \textbf{somatic operation} with $O(1)$ complexity. 

For a typical layer where $d$ might be 1024 or 2048, replacing $d$ multiplications with additive logic yields significant energy savings, as the cost of the single somatic scaling step ($1$ operation) becomes negligible compared to the eliminated synaptic workload ($d$ operations). Furthermore, as discussed in Section 3.1, this single scaling step can be reduced to a hardware-friendly bit-shift, ensuring that the entire pipeline remains efficient.

\begin{table}[h]
\centering
\caption{Comparison of Fundamental Operations per Perceptron (Input Dim: $d$)}
\label{tab:complexity}
\begin{tabular}{llcc}
\toprule
\textbf{Scope} & \textbf{Operation Type} & \textbf{Standard FFNN} & \textbf{NCN (Ours)} \\
\midrule
\multicolumn{4}{l}{\textit{\textbf{Synaptic Operations} (Scale: $O(d)$) - High Volume}} \\
& \textbf{Floating-Point Multiplication} & $\mathbf{d}$ & \textbf{0} \\
& Floating-Point Addition & $d$ & $2d$ \\
& Comparisons / Sign Checks & 0 & $2d$ \\
& Multiplexing (Mux) & 0 & $2d$ \\
\midrule
\multicolumn{4}{l}{\textit{\textbf{Somatic Operations} (Scale: $O(1)$) - Constant Time}} \\
& Bias Addition & 1 & 1 \\
& \textbf{Scaling / Normalization} & 0 & \textbf{1} \\
\bottomrule
\end{tabular}
\end{table}

\section{Experiments}

To validate the theoretical capabilities of Neuro-Channel Networks (NCN), we focused on fundamental non-linear logic problems. All experiments were conducted using a custom framework implemented in Python.

\subsection{Experimental Setup}
For all experiments, we used the same network structure defined in our framework. The network architecture consists of fully connected NCN layers followed by a standard Softmax layer for classification.
\begin{itemize}
    \item \textbf{Initialization:} Weights $w \sim \mathcal{N}(0, 1)$, Neurotransmitters $n \sim \mathcal{N}(0, 0.5)$.
    \item \textbf{Optimizer:} Standard SGD with Momentum ($0.9$).
    \item \textbf{Loss Function:} Cross-Entropy Loss.
\end{itemize}

\subsubsection{The XOR Problem}
The XOR (Exclusive OR) problem is a classic benchmark in the neural network literature. It requires the model to output 1 if the inputs are different $(0,1$  or $1,0)$ and $0$ if they are the same $(0,0$ or $1,1)$. This problem is historically significant because it proved that single layer perceptrons (which are linear classifiers) cannot solve non-linearly separable problems. In this experiment, although the forward pass is multiplication free, we utilized \textbf{standard backpropagation}~\cite{ref6} to update the channel widths, neurotransmitters, and biases.

\begin{table}[H]
\centering
\caption{Truth Table for the XOR Problem}
\label{tab:xor_truth_table}
\begin{tabular}{cc|c}
\toprule
\textbf{Input A} & \textbf{Input B} & \textbf{Output (Target)} \\
\midrule
0 & 0 & 0 \\
0 & 1 & 1 \\
1 & 0 & 1 \\
1 & 1 & 0 \\
\bottomrule
\end{tabular}
\end{table}

\begin{figure}[H]
    \centering
    \begin{tikzpicture}[x=1.5cm, y=1.5cm, >=stealth]
        \foreach \m [count=\y] in {1,2}
            \node [circle, draw, fill=green!20, minimum size=0.8cm] (input-\m) at (0, 2.5-\y) {$x_\m$};
        
        \foreach \m [count=\y] in {1,2,3,4}
            \node [circle, draw, fill=blue!20, minimum size=0.8cm] (hidden-\m) at (2, 3.5-\y) {$h_\m$};
            
        \foreach \m [count=\y] in {0,1}
            \node [circle, draw, fill=red!20, minimum size=0.8cm] (output-\m) at (4, 2.5-\y) {$y_\m$};
            
        \foreach \i in {1,2}
            \foreach \j in {1,2,3,4}
                \draw [->] (input-\i) -- (hidden-\j);
                
        \foreach \i in {1,2,3,4}
            \foreach \j in {0,1}
                \draw [->] (hidden-\i) -- (output-\j);
                
        \node [above, font=\bfseries] at (0, 2) {Input};
        \node [above, font=\bfseries] at (2, 3) {Hidden};
        \node [above, font=\bfseries] at (4, 2) {Output};
    \end{tikzpicture}
    \caption{Architecture of the NCN used for the XOR problem (2-4-2 topology).}
    \label{fig:xor_topo}
\end{figure}
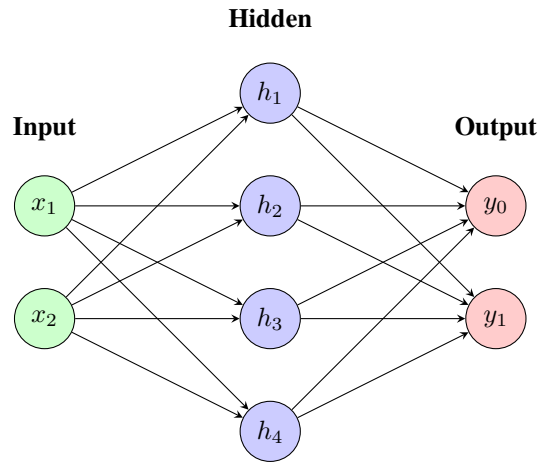

\textbf{Network Topology and Parameters:}
\begin{itemize}
    \item \textbf{Training Method:} Standard Backpropagation (SGD).
    \item \textbf{Topology:} $2 \to 4 \to 2$ (2 Input neurons, 4 Hidden neurons, 2 Output neurons). Shown in Figure~\ref{fig:xor_topo}.
    \item \textbf{Learning Rate:} $0.001$.
    \item \textbf{Momentum:} $0.9$.
    \item \textbf{Epochs:} 1000.
\end{itemize}

\begin{figure}[H]
    \centering
    \includegraphics[width=0.6\textwidth]{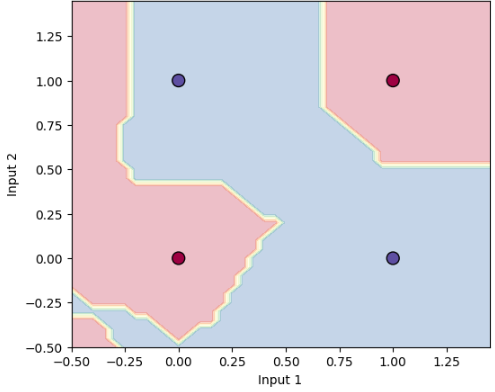}
    \caption{Decision boundary learned by NCN on the XOR function via standard backpropagation. The model successfully separates the non-linear classes using only additive operations.}
    \label{fig:xor}
\end{figure}

\textbf{Results:}
The NCN converged successfully using standard backpropagation, achieving \textbf{100\% accuracy}. The network topology used for this task is illustrated in Figure~\ref{fig:xor_topo}. Figure~\ref{fig:xor} illustrates the decision boundary learned by the network. As shown, the network successfully creates a non-linear separation, isolating the $(0,0)$ and $(1,1)$ regions from the $(0,1)$ and $(1,0)$ regions. This confirms that the interaction between the Channel Model utilized and Neurotransmitter Bypass creates effective non-linear transformations without needing multiplicative weights.

\subsubsection{The Majority Function}
We further tested the network on the $3-bit$ Majority function. This function outputs $1$ if the majority of the input bits (e.g., $2$ out of $3$) are $1$; otherwise, it outputs $0$. This tests the network's ability to perform threshold based logical reasoning and aggregate information from multiple sources. In this experiment, we again utilized \textbf{standard backpropagation}~\cite{ref6} to update the channel widths, neurotransmitters, and biases.

\begin{table}[H]
\centering
\caption{Truth Table for the 3-bit Majority Function}
\label{tab:majority_truth_table}
\begin{tabular}{ccc|c}
\toprule
\textbf{Input A} & \textbf{Input B} & \textbf{Input C} & \textbf{Output (Target)} \\
\midrule
0 & 0 & 0 & 0 \\
0 & 0 & 1 & 0 \\
0 & 1 & 0 & 0 \\
0 & 1 & 1 & 1 \\
1 & 0 & 0 & 0 \\
1 & 0 & 1 & 1 \\
1 & 1 & 0 & 1 \\
1 & 1 & 1 & 1 \\
\bottomrule
\end{tabular}
\end{table}

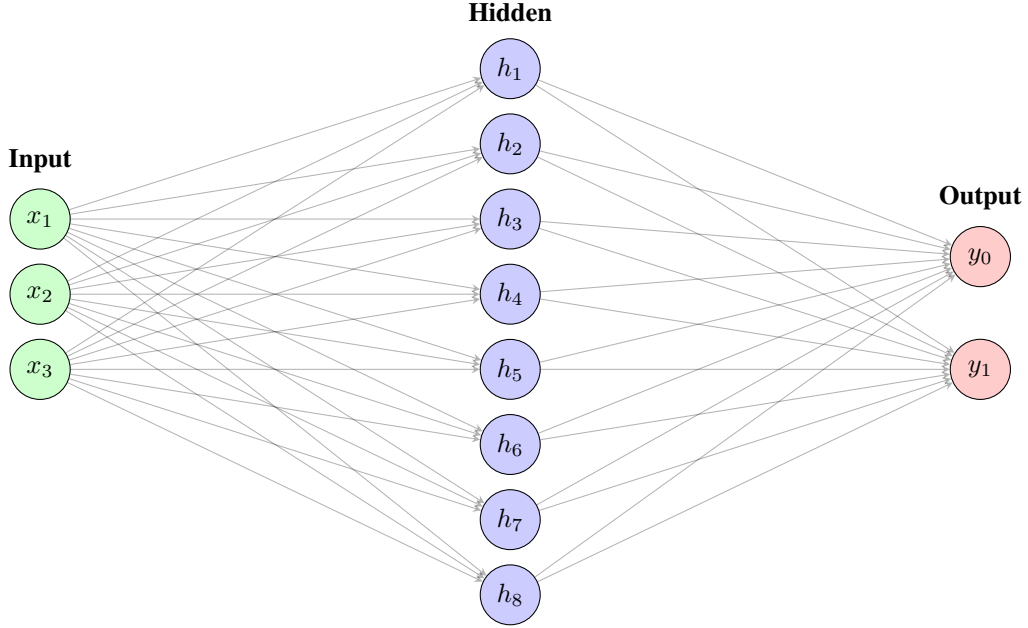
\begin{figure}[H]
    \centering
    \begin{tikzpicture}[x=2.5cm, y=0.5cm, >=stealth]
        \foreach \m [count=\y] in {1,2,3}
            \node [circle, draw, fill=green!20, minimum size=0.8cm] (input-\m) at (0, 5.5-\y*2) {$x_\m$};
        
        \foreach \m [count=\y] in {1,...,8}
            \node [circle, draw, fill=blue!20, minimum size=0.6cm] (hidden-\m) at (2.5, 9.5-\y*2) {$h_\m$};
            
        \foreach \m [count=\y] in {0,1}
            \node [circle, draw, fill=red!20, minimum size=0.8cm] (output-\m) at (5, 5.5-\y*3) {$y_\m$};
            
        \foreach \i in {1,2,3}
            \foreach \j in {1,...,8}
                \draw [->, opacity=0.3] (input-\i) -- (hidden-\j);
                
        \foreach \i in {1,...,8}
            \foreach \j in {0,1}
                \draw [->, opacity=0.3] (hidden-\i) -- (output-\j);

        \node [above, font=\bfseries] at (0, 4.5) {Input};
        \node [above, font=\bfseries] at (2.5, 8.5) {Hidden};
        \node [above, font=\bfseries] at (5, 3.5) {Output};

    \end{tikzpicture}
    \caption{Architecture of the NCN used for the 3-bit Majority function (3-8-2 topology).}
    \label{fig:majority_topo}
\end{figure}

\textbf{Network Topology and Parameters:}
\begin{itemize}
    \item \textbf{Topology:} $3 \to 8 \to 2$ (3 Input neurons, 8 Hidden neurons, 2 Output neurons). Shows in Figure~\ref{fig:majority_topo}.
    \item \textbf{Learning Rate:} $0.001$.
    \item \textbf{Momentum:} $0.9$.
    \item \textbf{Epochs:} 200.
\end{itemize}

\textbf{Results:}
The network successfully learned the complete truth table for all $2^3=8$ possible input combinations with \textbf{100\% accuracy}. This result demonstrates that NCNs can effectively aggregate multiple input signals and perform robust decision making.

\section{Discussion}

The results presented in the previous section demonstrate that the proposed Neuro-Channel Network (NCN) is capable of learning non-linear representations without relying on matrix multiplication in the forward pass. This finding has significant implications for both neuromorphic computing and hardware efficient AI.

\section{Conclusion and Future Directions}

In this paper, we introduced \textbf{Neuro-Channel Networks (NCN)}, a novel neural architecture inspired by the ion channel gating mechanisms of biological synapses. By replacing computationally expensive floating point multiplications with additive and bitwise logic operations, NCN challenges the prevailing arithmetic heavy paradigm of deep learning. Our empirical results on fundamental non-linear problems, such as the Majority and the XOR functions, establish the structural validity of the proposed \textbf{Signal Transmission} mechanism. We have demonstrated that complex, non-linear decision boundaries can be effectively constructed using only physical channel limits and neurotransmitter based gating, without utilizing multiplication in the forward pass.

Building upon this proof-of-concept, our future research will focus on three strategic pillars to bridge the gap between theoretical efficiency and practical deployment:

\begin{enumerate}

    \item \textbf{Implementation of Fully Multiplication-Free Training:} 
    While this study validated the forward pass efficiency, the training phase utilized standard backpropagation~\cite{ref6}. To achieve true end-to-end efficiency, our immediate future work will focus on extending the multiplication-free paradigm to the training phase. We aim to develop optimization strategies that replace standard gradient descent multiplications with additive or sign-based updates~\cite{ref15} (SignSGD), enabling the architecture to perform both inference and training using only bitwise operations and additions.
    
    \item \textbf{Scalability and Deep Representation Learning:} 
    We plan to extend the NCN architecture to deeper topologies and evaluate its capacity for hierarchical feature extraction on high-dimensional datasets such as MNIST and CIFAR-10. A key objective is to investigate whether the channel mechanism can approximate the performance of Convolutional Neural Networks (CNNs) in computer vision tasks while maintaining a significantly lower energy footprint.

    \item \textbf{Neuromorphic Hardware Co-Design:} 
    Theoretical complexity analysis suggests substantial energy savings. To quantify this, we intend to synthesize NCN logic onto Field Programmable Gate Arrays (FPGAs) and design custom Application Specific Integrated Circuits (ASICs). This will allow us to measure precise power consumption per inference.
\end{enumerate}

We believe that Neuro-Channel Networks represent a pivotal step towards biologically plausible artificial intelligence. By aligning mathematical operations with the physiological constraints of synapses, NCN offers a scalable pathway to overcome the energy bottlenecks of modern silicon, paving the way for the next generation of efficient and ubiquitous AI.


\begin{thebibliography}{15}

\bibitem{ref1}
Horowitz, M. (2014). 
``1.1 Computing's energy problem (and what we can do about it).'' 
\textit{IEEE International Solid-State Circuits Conference Digest of Technical Papers (ISSCC)}, 10-14.

\bibitem{ref2}
Courbariaux, M., Hubara, I., Soudry, D., El-Yaniv, R., \& Bengio, Y. (2016). 
``Binarized Neural Networks: Training Deep Neural Networks with Weights and Activations Constrained to +1 or -1.'' 
\textit{arXiv preprint arXiv:1602.02830}.

\bibitem{ref9}
Rastegari, M., Ordonez, V., Redmon, J., \& Farhadi, A. (2016).
``XNOR-Net: ImageNet Classification Using Binary Convolutional Neural Networks.''
\textit{European Conference on Computer Vision (ECCV)}, 525-542. Springer.

\bibitem{ref10}
Elhoushi, M., Chen, Z., Shafiq, F., Ye, Y. H., \& Zhang, Z. (2021).
``DeepShift: Towards Multiplication-Less Neural Networks with Power-of-Two Weights.''
\textit{arXiv preprint arXiv:1905.13298}.

\bibitem{ref3}
Chen, H., Wang, Y., Xu, C., Shi, B., Xu, C., Tian, Q., \& Xu, C. (2020). 
``AdderNet: Do We Really Need Multiplications in Deep Learning?'' 
\textit{Proceedings of the IEEE/CVF Conference on Computer Vision and Pattern Recognition (CVPR)}, 1468-1477.

\bibitem{ref11}
Maass, W. (1997).
``Networks of spiking neurons: The third generation of neural network models.''
\textit{Neural Networks}, 10(9), 1659-1671.

\bibitem{ref4}
Kandel, E. R., Schwartz, J. H., \& Jessell, T. M. (2000). 
\textit{Principles of Neural Science} (4th ed.). New York: McGraw-Hill.


\bibitem{ref6}
Rumelhart, D. E., Hinton, G. E., \& Williams, R. J. (1986).
``Learning representations by back-propagating errors.''
\textit{Nature}, 323(6088), 533-536

\bibitem{ref7}
LeCun, Y., Bottou, L., Orr, G. B., \& Müller, K. R. (2012). 
``Efficient backprop.'' 
\textit{Neural networks: Tricks of the trade} (pp. 9-48). Springer, Berlin, Heidelberg.

\bibitem{ref8}
Glorot, X., \& Bengio, Y. (2010).
``Understanding the difficulty of training deep feedforward neural networks.''
\textit{Proceedings of the thirteenth international conference on artificial intelligence and statistics}, 249-256.

\bibitem{ref12}
Turrigiano, G. G., Leslie, K. R., Desai, N. S., Rutherford, L. C., \& Nelson, S. B. (1998).
``Activity-dependent scaling of quantal amplitude in neocortical neurons.''
\textit{Nature}, 391(6670), 892-896.

\bibitem{ref13}
Greengard, P. (2001).
``The neurobiology of slow synaptic transmission.''
\textit{Science}, 294(5544), 1024-1030.

\bibitem{ref14}
He, K., Zhang, X., Ren, S., \& Sun, J. (2016).
``Deep residual learning for image recognition.''
\textit{Proceedings of the IEEE conference on computer vision and pattern recognition (CVPR)}, 770-778.

\bibitem{ref15}
Bernstein, J., Wang, Y. X., Azizzadenesheli, K., \& Anandkumar, A. (2018).
``signSGD: Compressed Optimisation for Non-Convex Problems.''
\textit{International Conference on Machine Learning (ICML)}, 560-569. PMLR.

\bibitem{ref16}
Rosenblatt, F. (1958).
``The perceptron: a probabilistic model for information storage and organization in the brain.''
\textit{Psychological review}, 65(6), 386.

\end{thebibliography}
\end{document}